# Sharing Behavior in Ride-hailing Trips: A Machine Learning Inference Approach

**Morteza Taiebat**[1,2], **Elham Amini**[3], **Ming Xu**[1,2,*]

[1] School for Environment & Sustainability, University of Michigan, Ann Arbor, MI, USA

[2] Department of Civil & Environmental Engineering, University of Michigan, Ann Arbor, MI, USA

[3] School of Information, University of Michigan, Ann Arbor, MI, USA

* Corresponding author (mingxu@umich.edu)

**Abstract:**

Ride sharing or pooling is important to mitigate negative externalities of ride-hailing such as increased congestion and environmental impacts. However, there lacks empirical evidence on what affect trip-level sharing behavior in ride-hailing. Using a novel dataset from all ride-hailing trips in Chicago in 2019, we show that the willingness of riders to request a shared ride has monotonically decreased from 27.0% to 12.8% throughout the year, while the trip volume and mileage have remained statistically unchanged. We find that the decline in sharing preference is due to an increased per-mile costs of shared trips and shifting shorter trips to solo. Using ensemble machine learning models, we find that the travel impedance variables (trip cost, distance, and duration) collectively contribute to the predictive power by 95% in the propensity to share and 91% in successful matching of a trip. Spatial and temporal attributes, sociodemographic, built environment, and transit supply variables do not entail significant predictive power at the trip level in presence of these travel impedance variables. Our findings shed light on sharing behavior in ride-hailing trips and can help devise strategies that increase shared ride-hailing.

**Highlights**

- Predict sharing behavior in Chicago's ride-haling trips using ensemble ML methods.
- Willingness to share a ride declined over 52% throughout 2019.
- Over time, per-mile cost of shared trips increased, shorter trips shifted to solo.
- Travel impedance variables have the highest predictive power in sharing behavior.

**Graphical Abstract**

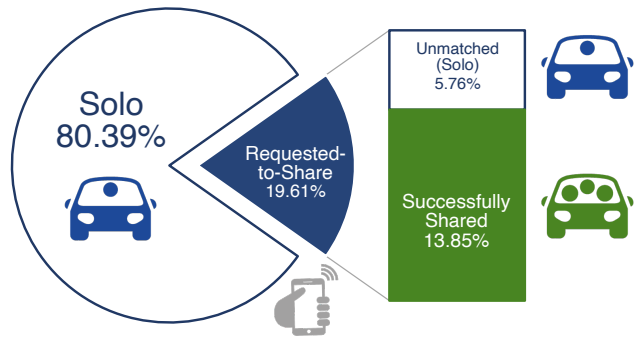

## 1. Introduction

Transportation Network Companies (TNCs) are rapidly transforming the urban and personal transportation. The increase in the adoption of ride-hailing (or ridesourcing) services such as Uber and Lyft can be attributed to the ease of access using a smartphone application along with a higher availability compared to the regulated, traditional taxi services (Conway et al., 2018; Cramer and Krueger, 2016; Schaller, 2018). Ride-hailing services offer flexible, efficient, and convenient mobility, promoted as a remedy for private vehicle dependency, traffic congestion, high parking costs, and environmental pollution. However, recent evidence reveals that the unintended consequences of ride-hailing services may outweigh some benefits by undermining public transportation (Grahn et al., 2020; Schwieterman and Smith, 2018), taking away from more sustainable transportation modes (Alemi et al., 2018; Clewlow and Mishra, 2017), increasing vehicle ownership, vehicle-miles-traveled (VMT) and large levels of deadheading miles (Henao and Marshall, 2019; Schaller, 2018; Ward et al., 2021, 2019; Wu and MacKenzie, 2021), and leading to aggravated congestion in urban areas (Erhardt et al., 2019). In large US ride-hailing markets, including metropolitan areas of Chicago, San Francisco, Washington, and Boston, it is estimated that 7-13% of total traffic in core counties is attributed to TNCs, while they serve only 2-3% of regional trips (Balding et al., 2019).

Ride sharing[1] or ride pooling, in which a rider shares all or portion of the trip with other passengers, has the potential to mitigate negative impacts of solo ride-hailing by consolidating VMT from multiple trips, which are spatially and temporally suitable for matching. The rate of sharing is a key factor in determining the sustainability of ride-hailing compared to other transportation alternatives, especially for future autonomous on-demand mobility services (Lavieri and Bhat, 2019a; Shaheen and Cohen, 2019; Taiebat et al., 2018). For instance, a number of recent studies found sharing a ride-hailing trip reduces GHG emissions by 10% to 47% (Anair et al., 2020; Li et al., 2021). Despite advantages of ride sharing (services such as UberPool and LyftShare), the portion of shared trips relative to solo trips in ride-hailing is still small (Schaller, 2021). In California, an average ride-hailing trip has 1.54 passengers, while an average household vehicle trip has 1.68 passengers (California Air Resources Board, 2019). Both city governments and TNCs are therefore trying to understand underlying factors of riders' willingness to share (WTS) their trips and factors of successful sharing. Nonetheless, lack of access to empirical data hinders better understanding of sharing behavior in ride-hailing trips. Currently, the vast majority of literature relies on revealed and stated preference surveys to understand ride-hailing adoption and usage (Alemi et al., 2018; Bansal et al., 2020; Chen et al., 2018; Lavieri and Bhat, 2019b; Wang et al., 2020). To address the data availability challenge and better understand and regulate ride-

---

[1] The standard and scientific terminology is ridepooling or ridesplitting, which is defined as a form of ridesourcing where riders with similar origins and destinations are matched to the same ride-hailing driver and vehicle in real-time, and the ride and costs are split among users (Shaheen et al., 2016, 2015; Shaheen and Cohen, 2019). Interchangeably, the term ride sharing is used in this manuscript with the same definition.

Morteza Taiebat, Elham Amini, and Ming Xu. "Sharing behavior in ride-hailing trips: A machine learning inference approach." *Transportation Research Part D: Transport and Environment* 103 (2022): 103166. DOI: 10.1016/j.trd.2021.103166hailing, many cities, including Chicago, implemented data mandates for active TNCs (Monahan, 2020).

Chicago is one of the largest ride-hailing markets in the US, where ride-hailing make up about 3% of the total regional VMT (Balding et al., 2019). The publicly available ride-hailing data from the City of Chicago has provided an unpreceded opportunity for empirical understanding of ride-hailing demand patterns (Ghaffar et al., 2020; Yan et al., 2020), relationship with transit services (Barajas and Brown, 2021), and neighborhood characteristics (Marquet, 2020). Since it has a unique feature of observing which trips were requested to be shared, recent studies with this data attempted to understand the determinants of WTS (Dean and Kockelman, 2021; Hou et al., 2020; Tu et al., 2021; Xu et al., 2021). The results of these studies reveal key factors affecting WTS, but only focusing on the aggregate level between origin-destination (O-D) pairs of census tracts. The WTS for individual trips, however, remains largely unknown. In addition, what factors determining the actual success of sharing once a rider requests to share the trip are also rarely studied. Understanding the trip-level determinants of WTS and sharing success is important for city governments and TNCs to design effective policies to encourage sharing in ride-hailing and improve the sustainability of urban transportation.

Building on the prior work, we examine this novel dataset of all ride-hailing trips in the city of Chicago (Chicago Department of Business Affairs & Consumer Protection, 2020) in 2019 to explore the factors affecting trip-level WTS (willingness of the rider to request a shared trip) and sharing success. By designing robust and generalizable ensemble machine learning (ML) prediction models, we find that the travel impedance variables (trip cost, length, duration) collectively contribute to 95% and 91% of the predictive power in determining the WTS of a trip and whether the trip that is requested to share is successfully shared or not, respectively. Specifically, for a dollar increase in the per-mile cost, we find about 58% reduction in the odds of a trip being requested to share. Unlike prior studies that found other variables (socioeconomic, demographic, built environment, spatial and temporal attribute, and transit supply variables) have high predictive power in determining the *portion* of trips requested to share between O-D pairs of census tracts, we find those variables do not entail predictive power in determining the WTS of individual trips. Our results imply that pricing signals have a higher potential to encourage riders to share their rides. We also find that longer, but less expensive requested-to-share trips are more likely to be successfully shared with another ride.

The reminder of the article is organized as follows. Section 2 reviews the literature and recent findings on factors that influence willingness to share the ride-hailing trips. Section 3 describes the dataset used in this study and provides an exploratory analysis of sharing behavior in ride-hailing trips in Chicago. Section 4 looks at the declining time-trend of trip sharing and attempts to understand the factors that influenced this decline, using a regression analysis. Section 5 describes the machine learning approach for prediction of willingness to share and successful sharing of ride-haling trips, and section 6 reports the results. Finally, section 7 provides a detailed discussion, draws conclusions on the findings, and explains major limitations of data and modeling approach used in this study that could thwart our inference. Our findings shed light on sharing behavior in ride-hailing trips and can help TNCs, urban planners, and policymakers to devise better strategies and targeted pricing mechanisms to increase sharing in ride-hailing. While sharing is currently suspended in most markets because of COVID-19, incentives can regain and improve the WTS of TNC users in the post-pandemic period, thereby helping avoid unintended congestion and environmental impacts from ride-hailing.

## 2. Factors associated with willingness to share

An extensive body of travel behavior literature is dedicated to understanding the theoretical aspects as well as stated and releveled preference for sharing the ride and sharing for ride-hailing. These studies predominantly use surveys combined with the neoclassical econometric models to understand the preference for using ride-hailing services in general, as well as investigating the factors that contribute to riders opting for a shared ride-hail trip (Alemi et al., 2018; Clewlow and Mishra, 2017; Kang et al., 2021). A majority of these stated-preference studies suggest that employed, educated, urban residents living in high density, walkable neighborhoods are more likely to share rides and use ride-hailing in general than those hailing rides from predominately non-white, older, or low-income neighborhoods. They also state that underlying discomfort with sharing a ride with strangers and increase of travel time are among major barriers of sharing (Alonso-González et al., 2020b, 2020a; Bansal et al., 2020; Kang et al., 2021; Li et al., 2020; Wang et al., 2020). Shared rides can take longer than private ride-hailing trips due to time and mileage penalties associated with detours for pickups and drop-offs of the matched rides (Wang et al., 2020; Young et al., 2020). Therefore, such trips might not be a viable option when traveling to time-sensitive appointments (work, doctor's visit, airport, etc.) or for riders who place a high value on travel time. With shared rides costing less than regular trips, individuals would have to weigh the cost savings against increased travel times, as well as discomfort (if any) in traveling with strangers.

Converging evidence from revealed preference and empirical data from sharing behavior is scarce, and in some cases contradictory to the previous understanding (Li et al., 2019). Using trip data in Los Angeles, Brown found riders living in low-income dense areas make higher proportion of shared trip, and 10% of riders make 94% of shared trips (Brown, 2020). Young et al. found that higher demand and longer trip distances significantly improve matching propensity for shared ride-hailing trips in Toronto (Young et al., 2020). Tu et al. found that distance to city center, land use diversity and road density are the key influencing factors of sharing behavior (Tu et al., 2021). Leveraging the same Chicago dataset used in our study, Hou et al. and Xu et al. took a similar approach to study the ratio of shared trips to total trips between O-D pairs (binned by pickup and drop-off census tracts) as a regression-based ML problem (Hou et al., 2020; Xu et al., 2021). They found that that socio-demographic variables as well as pickup/drop-off in airport census tracts have the highest predictive power, but both reported relatively large unexplained variance and high sensitivity to outlier observations. Dean and Kockelman employed a more nuanced econometric approach to model the count and ratio of shared ride trips. They found the spatial accessibility variables and underlying socioeconomic characteristics of the origin zones significantly influence the proportion and count of shared ride-hailing trips (Dean and Kockelman, 2021). All these studies reveal the association of requested-to-share trips with explanatory variables. However, there is less empirical evidence on the factors that influence successfully shared trips, which are a subset of requested-to-share trips.

## 3. Ride-hailing trip data and exploratory analysis

The City of Chicago has had a data sharing mandate in place for all active TNCs (Lyft, Uber, and Via) since November 2018, as a part of compliance and operation licensing framework. We use the city's person-trip level database, which records, most importantly, when and where a ride (that is, person-trip) happens and whether the rider (a) "requested to share" the ride and (b) successfully shared the ride. The raw data for this study is available from the City of Chicago data portal (Chicago Department of Business Affairs & Consumer Protection, 2020). This spatiotemporal data is at the trip level and contains trip attributes including ID, start/end timestamps, duration, distance, pickup/drop-off census tracts, fare, additional charges, tip, and trip total cost. Every trip record has an indictor to show whether the rider requested sharing (i.e., whether the rider is willing to take a potential shared trip). Each trip also has a binary indictor to show whether they were successfully shared or not. For simplicity, we henceforth define the trips that are requested (or authorized) to share as "requested-to-share trips" and the subset of requested-to-share trips that are successfully shared as "shared trips". Note that the data does not directly measure or represent willingness to share/pool, and we loosely relate it (WTS) to requested-to-share trips as a proxy here, consistent with prior studies of Dean and Kockelman, 2021 and Hou et al., 2020). Nonetheless, behavioral aspects, willingness and attitudes toward sharing are measured through surveys.

The City of Chicago has applied de-identification and aggregation techniques to reduce the risk of linking individuals' trip data to their identities. This includes aggregating the pickup and drop-off locations at the census tract level[2], rounding the trip-start and trip-end times to the nearest 15 minutes, and rounding the fares and tips to the nearest $2.50 and $1.00, respectively. Nearly 16.5% of trips suffer from missing pickup and/or drop-off census tracts, 77% of which represent the trips that requested sharing. The data provider cites privacy concerns in masking these values. However, the de-identification process may have impacted the data quality. Specifically, rounding the fares may induce bias in our inference.

For this study, we focus on data from the entire year of 2019 (111.85 million trip records before cleaning). Since the pickup and drop-off are reported at the census tracts, we can augment many explanatory variables from auxiliary datasets at the census tract level. We derive spatial and temporal attributes of trips from the main Chicago dataset. We also append the trip data with three sets of variables from auxiliary datasets: 1) socioeconomic and demographics variables from American Community Survey (ACS) (U.S. Census Bureau, 2020) and Chicago Metropolitan Agency for Planning (CMAP) (Chicago Metropolitan Agency for Planning); 2) built environment variables from ACS, CMAP, Longitudinal Employer-Household Dynamics (LEHD) (U.S. Census Bureau., 2020), and Google Map API; and 3) transit supply variables from General Transit Feed Specification (GTFS). The census tract level data from auxiliary datasets is provided in the Supporting Information.

---

[2] Although the pickup and drop-off locations are suppressed to the centers of census tracts, the trip distance and duration have not been impacted by the de-identification process and represent the real mileage and real time interval (in seconds) of the trip, respectively.

The data requires significant cleaning effort as well as dealing with significant portion of missing data. For the procedure for data cleaning and imputation of missing values, we first remove all incorrect observations which can be characterized as inconceivable trips.[3] To protect the privacy, the pickup and/or drop-off of some trips are masked (missing). The majority of missingness is from the requested-to-share trips and census tracts in outskirts. Since this missingness is not at random, simply removing those observations changes the distribution of requested-to-share trips relative to all trips and also hinders their spatial variance. This could bias our analysis and thus, we follow an imputation strategy to overcome the non-random missingness. We first attempt to infer the pickup or drop-off tracts from the pickup or drop-off community code if it is not missing. The City of Chicago has 77 community areas and within each community area there are multiple census tracts. We group the dataset by community area and impute missing census tracts within each community area by trip-density weighted ranking of the non-missing census tracts in that group. This imputation reduces missing values to 5.8% but may induce a modest bias in our estimation (note that Xu et al., 2021 used a comparable strategy for imputation). The data also includes some trips outside the boundaries of the City of Chicago (Cook County). We also remove census tracts outside of the City of Chicago based on the 2010 census boundaries (801 census tracts). After the cleaning and imputation processes, 12.1% of trips were removed, bringing down the total number of observations to 96,268,064 trip records for all 12 months in 2019 covering over 470 million miles.

---

[3] The likely inconceivable trips include trips with total trip duration less than 1 minute and longer than 5 hours; trips less total distance traveled than 0.25 miles and greater than 300 miles; trips with total fare equal to zero (fares are already rounded); trips with extreme speeds (below 0.2 mph and above 80 mph - an auxiliary variable from trip distance and trip duration). We removed all of these trips from the dataset.

We observe that from over 96 million trips in 2019, on average, only 19.6% of trips were requested-to-share trips and less than 70% of those were successfully matched with another ride, thus only 13.8% of trips were shared. Fig. 1 provides a detailed exploratory analysis of the data, showing spatial and temporal variation of requested-to-share and shared trips. The average portion of shared requested trips and successfully shared trips based on hour of the day and day of the week has a relatively similar pattern to trip demand with some distinct differences. The peak hours of demand for TNCs in Chicago are 7AM-9AM and 5PM-7PM weekdays, 10AM-12AM and 5PM-7PM on weekends. As shown in Fig. 1A, the average share of requested-to-share trips increases during the weekday peak demand hours, likely due to the fact that commuter trips during these hours are more likely to share (Kang et al., 2021). It also peaks at midnight reaching over 25% except for Friday and Saturday. The share of shared trips follows a similar pattern, which is expected as more requested trips naturally yield more shared trips. However, the portion of successfully shared trips drops precipitously after the midnight, despite the portion of requested-to-share trips remains relatively high. This drop is likely due to lower supply of ride-hailing vehicle overnight.

Fig. 1B shows the share of trips and requested-to-share trips in different origin and destination areas including downtown, airport, economically disconnected areas (EDA), and other census tracts. Chicago has two major airports, and the downtown zone consists of 30 census tracts. EDA consists of 489 tracts with higher than regional average concentrations of low-income households and minorities (Chicago Metropolitan Agency for Planning). We observe that in these areas the rate of requested-to-share trips is significantly higher than the rest of the city. Among those, nearly 40% of EDA-to-EDA trips are requested to be shared (twice the city-wide average). Looking at the average spatial variation, as shown in Fig.2, we can observe that the downtown and airport trips are associated with the low average rates of shared requested trips and successfully shared trips, while census tracts associated with lower income residents and higher percentage of African American residents (majority in EDA) have higher level of requested-to-share and shared trips.

Fig. 3 shows the trend of requested-to-share trips versus solo-trips in EDA and non-EDA. Trips with either or both pickup and drop-off in 489 EDA census tracts are labeled as EDA trips. The rate of requested-to-share trips in EDA is twice as non-EDA trips. EAD trips are consistently longer and more expansive for both solo and requested-to-share trips compared to non-EDA trips. Fig. S-2 and Fig. S-3 show the kernel density distribution of key variables for requested-to-share and successfully shared trips. The distributions of the total cost, distance, duration, and per-mile cost of requested-to-share and solo trips, as well as successfully shared and unmatched subset of requested-to-share trips, significantly overlap, despite that the mean differences are statistically significant (mainly due to very large sample size). Due to frequent surge pricing that TNCs impose on trips (for instance, when vehicle supply is lower than trip demand), the cost per mile or time of a requested-to-share trip could be higher than a solo one from the same pick-up and drop-off, depending on the time of the day and many other demand factors that are not observable.

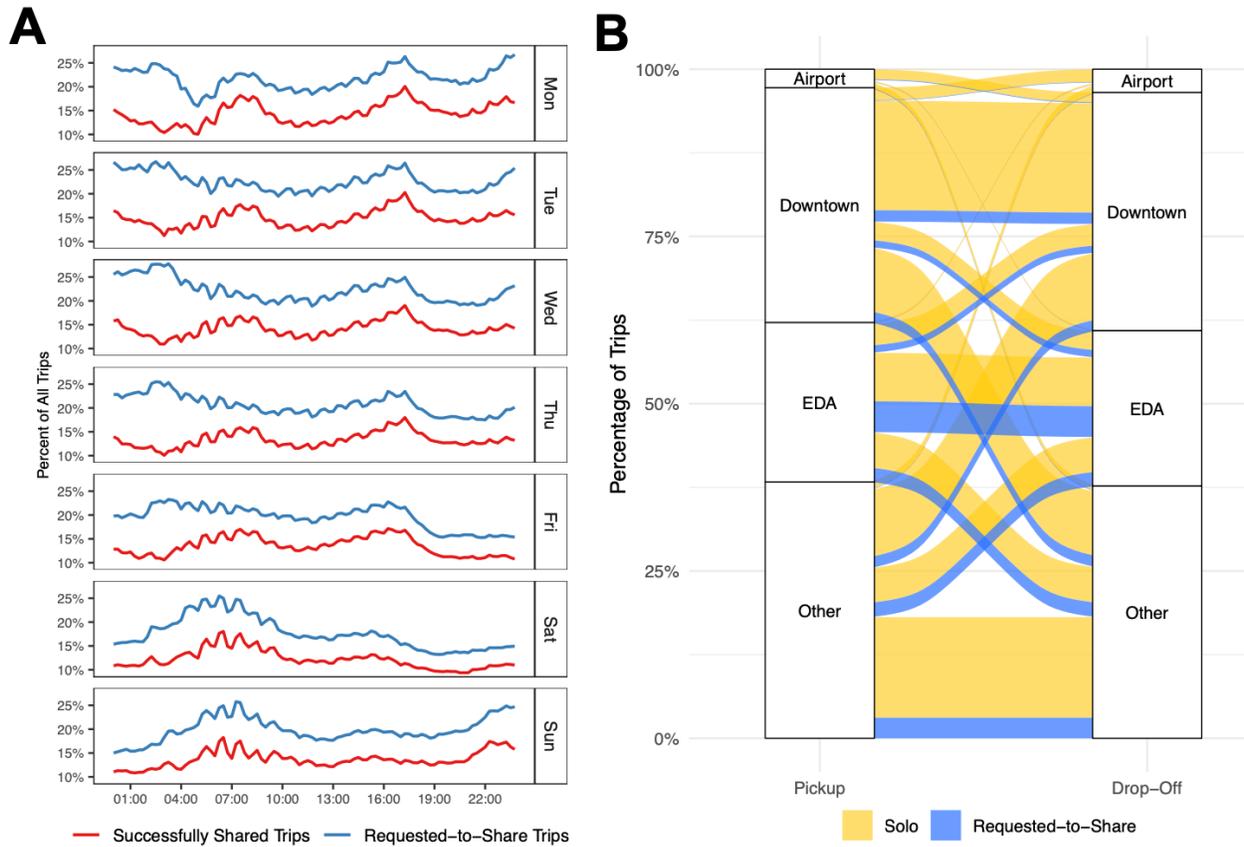

Fig. 1. Exploratory analysis of requested-to-share trips and successfully shared trips. (A) Day of the week and time of day average rates of requested-to-share and successfully shared trips (15-minute window average). (B) Flow of trips between downtown, airport, economically disconnected areas (EDAs) and other census tracts of Chicago as pickup and drop-off points. The portion of requested-to-share trips is show in blue.

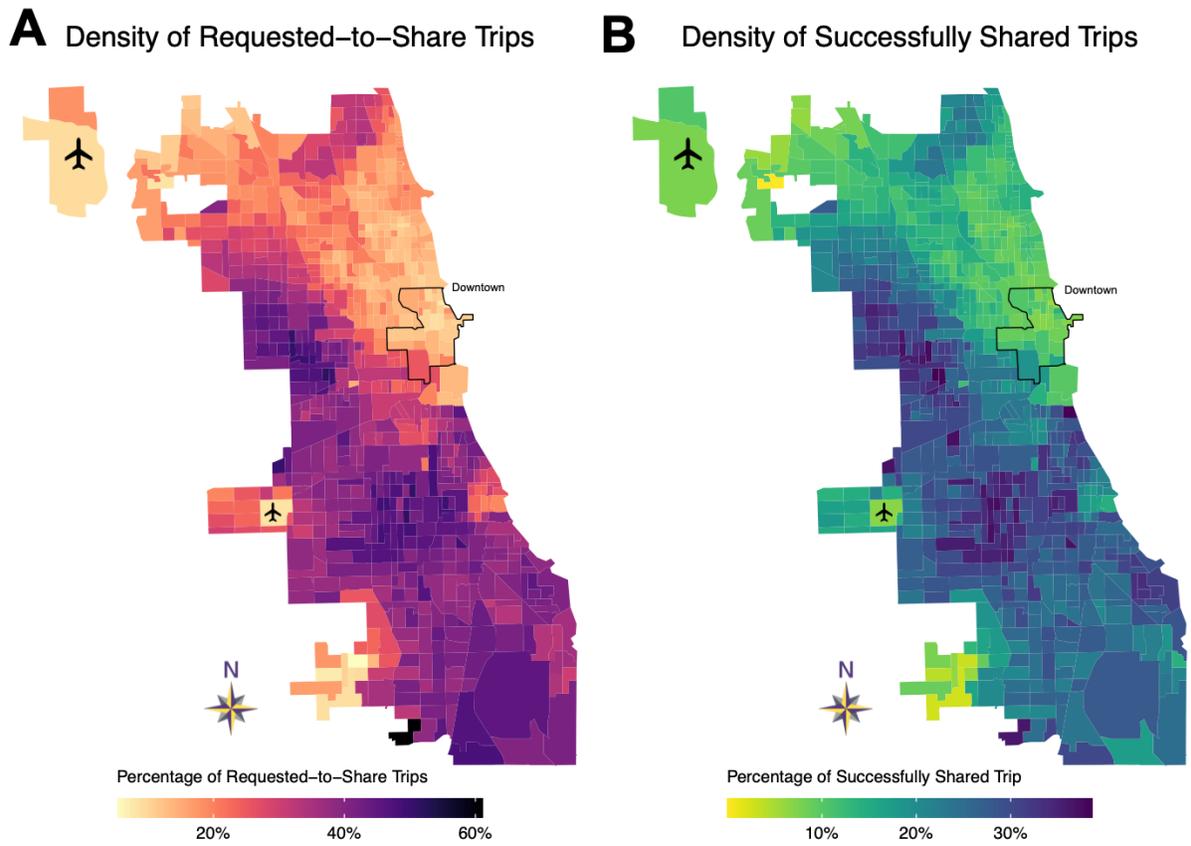

Fig. 2. Density of requested-to-share and successfully shared trips by the pickup census tract. The downtown area is shown in black line. There densities with the drop-off census tract are highly correlated with those of pickup census tract.

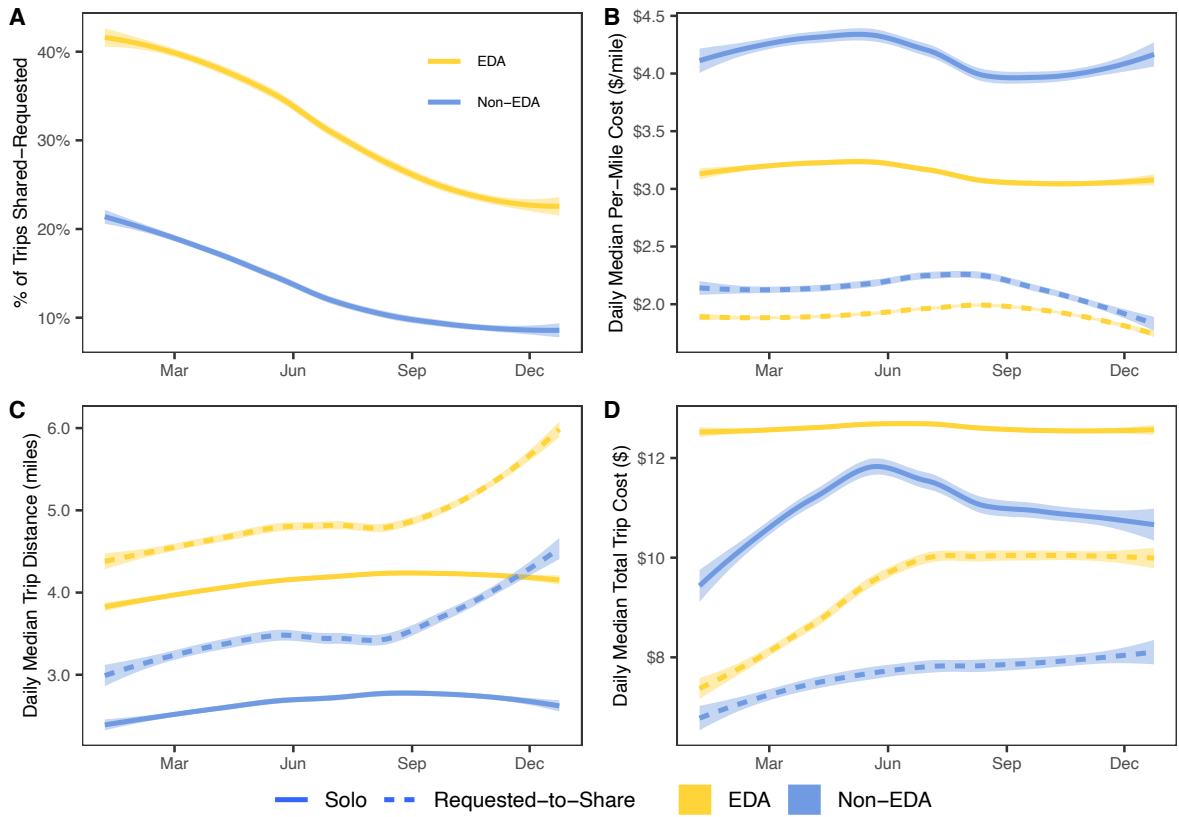

Fig. 3. The smoothed trend of requested-to-share trips versus solo-trips in EDA and non-EDA. (A) The percentage of requested-to-share trips among all trips; (B) Daily median per-mile cost of the trip; (C) Daily median trip distance, (D) Daily median total cost of the trip.

## 4. Trends of sharing in Chicago's ride-hailing trips

Fig. 4A shows that riders' WTS declined throughout the year 2019, confirmed by a Mann-Kendall test showing monotonic downward trend of the portion of requested-to-share trips ($\tau = -0.89$, $P < 10^{-4}$) which decreased from 27.0% in January to 12.8% in December. However, the trip volume and mileage have remained statistically unchanged in this period. The portion of shared trips closely follows the same trend with a shift, indicating a relatively stable rate of successful sharing—67-71% of requested trips were actually shared throughout the year. Intuitively, a lower demand for requested-to-share trips would further reduce the successful sharing, since the matching algorithm would have fewer potential rides to choose/match with. This counterintuitive observation here, where the rate of successful sharing remained stable, may be explained by an unobserved systematic factor. For instance, the matching algorithms continuously maintain a metric by design such as successful matching rate.

As show in Fig. 4B, the continuous decline of WTS may be because the increasing cost of requested-to-share trips over time. Specifically, the median cost of a requested-to-share trip increased substantially in 2019, while the median cost of solo trips remained relatively stable throughout the year. The reduced incentive to share the ride due to higher cost might explain why more travelers opted for solo rides over time. However, the per-mile cost of the requested-to-share and solo trips (dotted lines) was relatively stable with median of $1.99/mile and $3.76/mile, respectively. The per-mile cost of requested-to-share trip even decreased in the second half of 2019 while the per-trip cost continued to increase. This indicates that riders' WTS for longer trips increased over time while that for shorter ones declined (Fig. S-1).

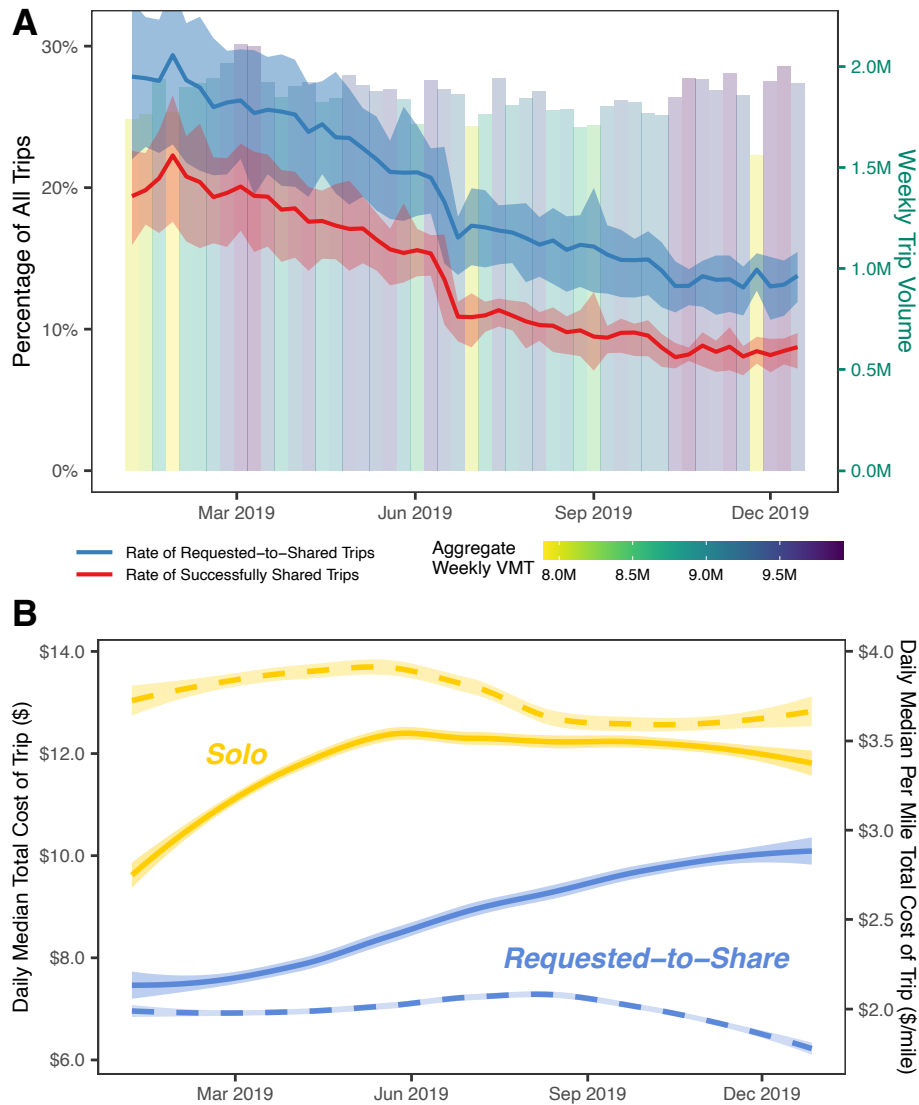

Fig. 4. Trend of sharing behavior in 2019. (A) Weekly average rates of requested-to-share rides, successful sharing, weekly trip volume and aggregate VMT. Shaded areas show daily variations. (B) The solid lines represent the smoothed daily median cost of requested-to-share trips and solo trips (left y-axis). The dashed lines represent the smoothed daily median per-mile cost of requested-to-share trips and solo trips (right y-axis).

To understand how trip attributes might lead to the decline of WTS over time, we run a logistic regression where the covariates are interacted with time (week of the year) as a continuous variable. This estimator explicitly examines the marginal effect of time by adjusting the response with the main effects. A general form of $logit(P) = \beta_0 + \beta X + \beta' X \times week + \varepsilon$ is employed where $P$ is the probability of requesting a shared ride, $logit(P)$ representing the natural log transformation of odds, $X$ is a vector of trip characteristics, and $\varepsilon$ is the error term. In this specification, $\beta + \beta' week$ represents the time-varying marginal effect of a covariate. We take a random stratified sample of 100,000 trip observations from the dataset to fit the model. Since trips are collapsed at the pickup and drop-off census tract levels, there is a possibility of correlation within the tract clusters. Thus, we cluster-adjust the standard errors within each pickup census tract to account for correlated unobserved components in outcomes for trips within tracts.

As shown in Table 1, all main effects are statistically significant. As expected, a unitary increase in time (week) reduces the WTS by 4% (more precisely, the odds of a trip being requested to share). Unitary increase in per-mile cost of the trip reduces the odds of WTS by 58%. Having one or both legs of the trip in the downtown area reduces the WTS, and a trip that started or ended in EDAs increases the WTS. While the main effects are important determinants of WTS, the decline of WTS over time can be explained by the estimated coefficients for time interaction terms. The per-mile cost and trip distance both have positive signs when interacted with time. Given that higher per-mile cost and longer distance without time interactions reduce the odds of WTS with statistical significance, the preference of riders to share shorter rides declined over time and a greater number of shorter trips were requested solo. The interactions of time with downtown and EDA indicators are not significant.

Table 1. Logistic regression results to understand the time trend of sharing behavior

| Covariate | Parameter estimates | | Odds ratio | | |
|---|---|---|---|---|---|
| | Parameter | Clu. Std. Err. | Estimate | LB (95% CI) | UB (95% CI) |
| *Main Effects* | | | | | |
| Week of year | -0.0448*** | 0.006 | 0.9561 | 0.9441 | 0.9683 |
| Per-mile cost | -0.8614*** | 0.0411 | 0.4225 | 0.3898 | 0.4580 |
| Trip distance | -0.2019*** | 0.0133 | 0.8171 | 0.7959 | 0.8389 |
| Downtown indicator | -0.3381*** | 0.0497 | 0.7131 | 0.6469 | 0.7860 |
| EDA indictor | 0.3077*** | 0.0516 | 1.3603 | 1.2294 | 1.5051 |
| *Time Interactions* | | | | | |
| Week × Per-mile cost | 0.0039** | 0.0014 | 1.0039 | 1.0012 | 1.0067 |
| Week × Trip distance | 0.0032*** | 0.0003 | 1.0032 | 1.0025 | 1.0038 |
| Week × Downtown Ind. | -0.0021 | 0.0013 | 0.9979 | 0.9953 | 1.0006 |
| Week × EDA Ind. | 0.0016 | 0.0014 | 1.0016 | 0.9986 | 1.0045 |
| Constant | 3.4182*** | 0.2078 | | | |

Notes:
- Solo trip is a base category and parameters are estimated for requested-to-share trip.
- Sample size: 100,000; Loglikelihood: -77937.86; Pseudo R-square: 0.188; AIC: 78,058.
- LB (95% CI) and UB (95% CI) imply lower and upper bounds of 95% confidence interval.
- Standard errors are clustered by pickup census tract (Clu. Std. Err.). All variables in the final model have variance inflation factor (*VIF*) < 10.
- Asterisks denote 1 (***), 5 (**), and 10 (*) percent significance levels.
- EDA indicator denotes whether the trip had a leg in economically disconnected areas (EDA).
- Other variables included in the regression are indictors for whether the trips had a leg in white majority (negative***) census tract, black majority census tract (positive***), airport (negative***), and normalized median income of pickup tract (negative***). The first three variables are also interacted with week, but the coefficients are not statistically significant at any level.

## 5. Predicting willingness to share and successful sharing

To predict the probabilities of a trip being requested to be shared and a requested-to-share trip being successfully shared, we use several ensemble ML methods and compare the performance. The goal of ensemble learning is to combine decisions or predictions of several weak classifiers to build non-parametric and interpretable predictive models to improve prediction, generalizability, and robustness over a single classifier (Hastie et al., 2009). We utilize Python-based Scikit-Learn implementation of Adaptive Boosting or AdaBoost (Ada) (Freund and Schapire, 1997), Gradient Boosting (GB) (Friedman, 2001), and Random Forests (RF) (Breiman, 2001). These popular ML algorithm use decision trees as base learners but operate differently in ensembling the trees. RF trains multiple decision trees in parallel on bootstrapped samples of training data and combines the predictions of all the decision trees to generate a final prediction. Boosting ensemble methods consist of fitting several weak learners sequentially, where, at each learning iteration, more weight is added to the observations with the worst prediction from the previous iteration. GB and Ada use different loss functions and the former is more robust to outliers than the latter. Interested reader are referred to (Hastie et al., 2009) for a more detailed introduction of each algorithm.

In addition to their ability to capture nonlinear relationships, these ensemble models deal well with both numerical and categorical variables and are robust to such issues as feature multi-collinearity, imbalanced datasets, and the existence of outliers and missing values (Hastie et al., 2009). The attributes of selected ensemble classifiers are particularly valuable given the characteristics of our dataset, including class imbalance and high dimensionality (large set of exploratory variables). The interpretability of models enables us to understand the relationship between the input variables and the prediction. The framework of our classification is depicted in Fig. 5.

After data preparation, building and training our model involved several steps: hyperparameter tuning, feature selection, model validation, calibration, and interpretation. These steps are explained in Appendix. The performance of models on the test set is evaluated using area under the receiver operating characteristic curve (ROC-AUC), prediction accuracy, precision, and recall as described in the performance metrics section.

Informed by the literature and our exploratory analysis, we choose six categories of explanatory variables from the main dataset and auxiliary data sources to explain sharing behavior. The travel impedance variables and spatial and temporal attributes of trips are at the trip level. The socioeconomic, demographic, built environment, and transit supply variables are at the census tract level according to pickup and drop-off locations. Each trip is described by 45 explanatory variables (features), as shown in Fig. 5. The variables of interest (target response) are binary and reflect whether the trip is requested to be shared and whether a requested-to-share trip is successfully shared. The level of correlation between binary targets and selected features are shown in Fig. S-4 and Fig. S-5. We only select explanatory variables which have a variance inflation factor (VIF) score below the threshold of 10, in order to avoid multi-collinearity issues in the ML process (Hastie et al., 2009).

Using learning curves with different sample sizes of input data (Fig. S-6), we find that with 8,000 data points our models fully saturate. Thus, to reduce training time, we use a randomly chosen stratified sample with 8,000 observations. Although it is a small fraction of the entire dataset of 96 million trip records, feeding the models with more data does not improve the performance anymore. As a robustness check and to quantify the performance uncertainty, we repeat the entire learning process with 100 bootstrapped random samples size of 8,000 from the entire dataset. Note that the random sampling procedure does not substitute the imputation procedure that was explained before. Removing the observations with non-random missingness of pickup and/or drop-off is not appropriate, since it changes the distributional balance of the data and biases these models.

We split the stratified random learning sample of 8,000 trips to 75% training and cross-validation, and 25% testing (hold-out-sample). All models are evaluated using suitable performance metrics for classification, including accuracy, precision, and recall on the testing set. The accuracy is the proportion of correct predictions. The precision evaluates the fraction of correct classified instances among the ones classified as positive, and the recall (sensitivity) quantifies the number of correct positive predictions made out of all positive predictions that could have been made. We evaluate the overall performance of a classifier by the area under the receiver operating characteristic curve (AUC-ROC).

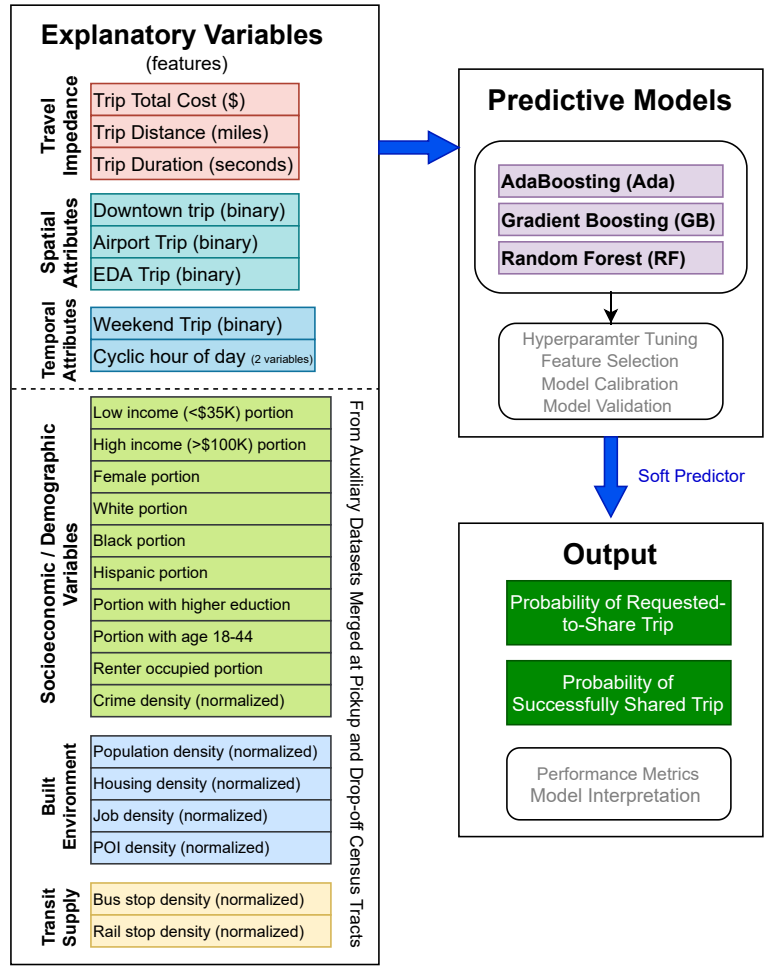

Fig. 5. ML framework to predict requested-to-share trips and successfully shared trips.

## 6. Results

The learning process reveals the variables with the highest predictive power. We measure the predictive power of a variable by iterative measuring of how model performance decreases when a variable is permutated (Breiman, 2001). We prefer permutation feature importance to impurity-based measures given its robustness to inflating the importance of numerical features, which may overfit the model. We find the travel impedance variables (trip cost, distance, and duration) have the highest predictive powers for both WTS and sharing success. Fig. 6A shows that for the requested-to-share trip prediction, the trip cost has the highest predictive power in all three models (Ada, RF, GB) and the three travel impedance variables collectively represent 89%-95% of predictive power of models. For prediction of successful sharing, Fig. 6B reveals that the travel duration entails the highest relative variable importance. Other variables including socioeconomic, demographic, spatial and temporal attribute, built environment, and transit supply variables at the origin and destination census tract of the trip show trivial predictive power compared to the travel impedance variables.

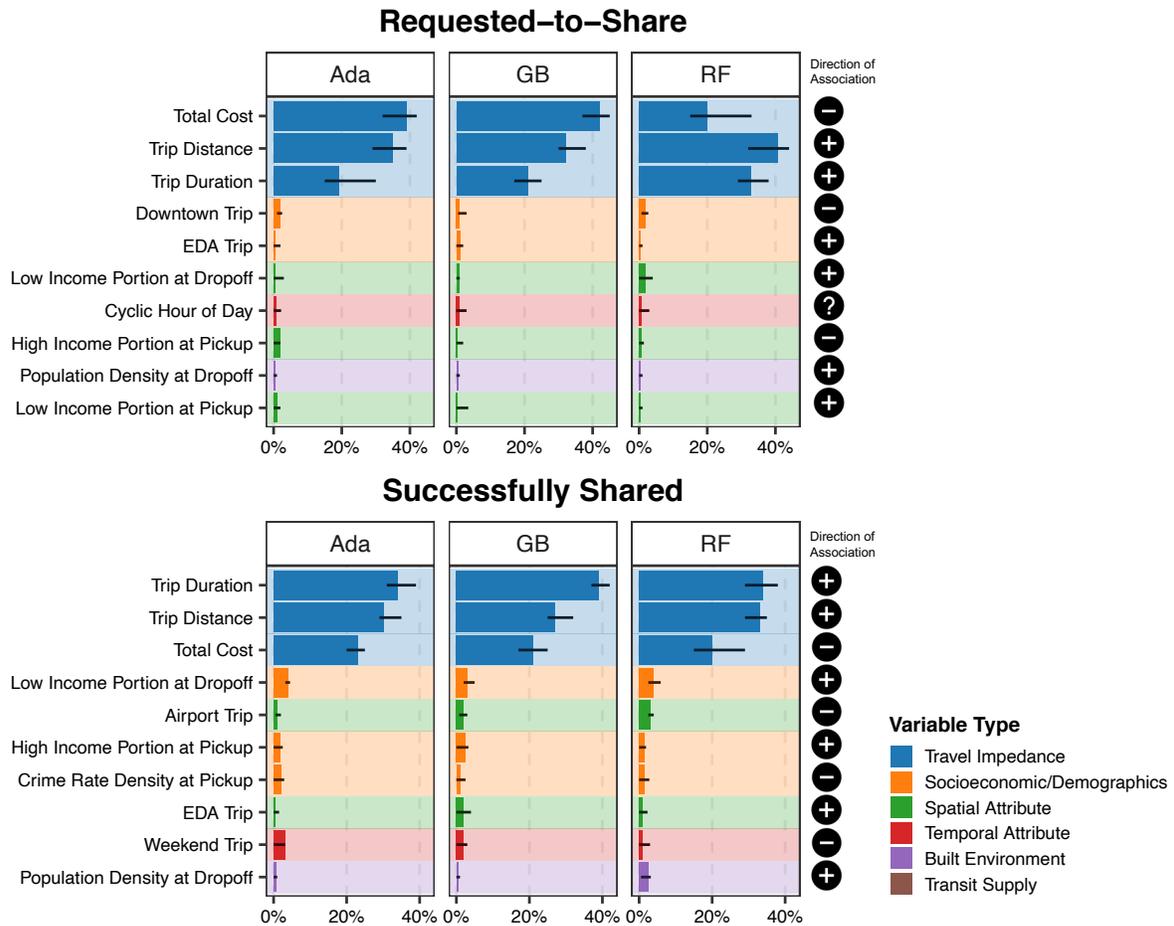

Fig. 6. Normalized relative predictive power of variables based on permutation feature importance with 5-fold cross-validation. Only top ten variables with the highest median predictive power are shown. The direction of association for each variable is determined using partial dependence analysis.

We use the results of permutation feature importance to choose the top ten relevant variables for the training and fine-tuning the classifiers. Irrelevant features simply add noise to the training data and affect the classification accuracy (Hastie et al., 2009). For example, adding noisy features such as transit supply variables increases the classification error. After hyperparameter tuning, the performance of Ada, RF, and GB classifiers are evaluated using several metrics. To show the overall performance of different classifiers for prediction of requested-to-share trips and successfully shared trips, ROC curves are illustrated in Fig. 7. Specifically, the Ada and GB models outperform the RF model for predication of requested-to-share trips even after significant hyperparameter tuning of the RF model. However, all three classifiers have a comparable performance in predicting successful sharing. We find that the shard-requested predictive models show higher performance than successful-sharing predictive models. This is likely due to more stochastic nature of successful matching affected by availability of nearby trips, traffic conditions, and the performance of TNC dispatch algorithms.

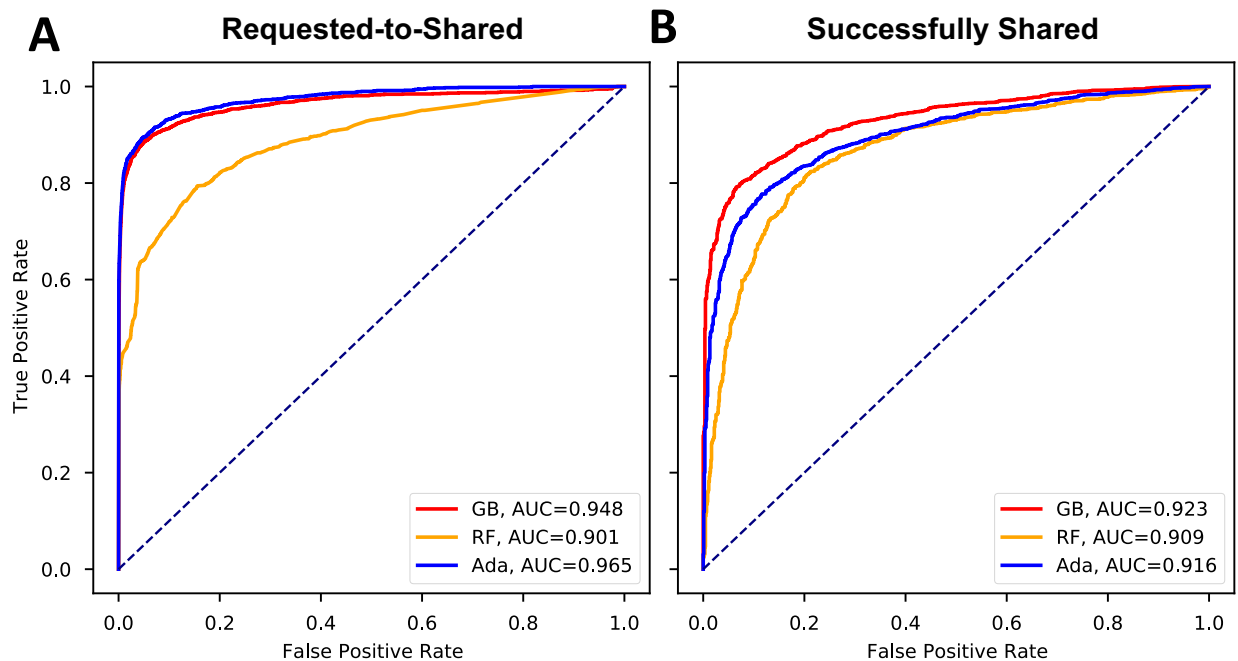

Fig. 7. Performance of ML models in prediction of (A) requested-to-share trips and (B) successfully shared trips. The area under the receiver operating characteristic (AUC-ROC) curve is equivalent to the probability that the model will rank a randomly chosen positive instance higher than a randomly chosen negative instance. The higher the AUC, the better a classification model. The diagonal line represents the baseline null classifier.

Table 2 provides more details on the performance metrics of final optimized classifiers. Since recall and precision are more important than accuracy in this problem, we specifically tune the classifiers' hyperparameters to maximize recall instead of accuracy. In this process, the RF model trades off a significant improvement in recall and precision to a slight decline in accuracy compared to the untuned model. This is mainly due to aggressive overfitting nature of the RF method before tuning (Fig. S-6). The Ada model shows superior classification performance with the highest recall rate for predicting requested-to-share trips while maintaining over 96% accuracy. For prediction of successfully shared trips, all three models are reasonably accurate, but the GB model maintains a slightly higher edge in other performance metrics.

Table 2. Performance of hyperparameter-tuned models on the test set. The percentage performance improvement compared to the untuned model on the validation set is shown in parentheses.

| Models | Prediction of Requested-to-Share | | | Prediction of Successfully Shared | | |
|---|---|---|---|---|---|---|
| | *Accuracy* | *Precision* | *Recall* | *Accuracy* | *Precision* | *Recall* |
| Random Forest | 0.92 (-1%) | 0.93 (-2%) | 0.74 (+4%) | 0.89 (-1%) | 0.91 (-1%) | 0.91 (+3) |
| Ada-Boosting | 0.96 (+3%) | 0.94 (-1%) | 0.82 (+6%) | 0.90 (0%) | 0.92 (+1) | 0.90 (+1%) |
| Gradient Boosting | 0.95 (+1%) | 0.94 (0%) | 0.80 (+4%) | 0.91 (+1%) | 0.91 (+2%) | 0.92 (+1%) |

To assess the relationship between the target responses of models (WTS and probability of sharing success) and selected explanatory variables, we use partial dependence plots (PDP). They intuitively show the marginal effect that specific features have on the predicted outcome of a model (Friedman, 2001; Goldstein et al., 2015). The PDPs for top three variables are shown in Fig.

Morteza Taiebat, Elham Amini, and Ming Xu. "Sharing behavior in ride-hailing trips: A machine learning inference approach." *Transportation Research Part D: Transport and Environment* 103 (2022): 103166. DOI: 10.1016/j.trd.2021.103166S-7 and Fig. S-8, capturing the highly nonlinear relationship with the target responses. The direction of associations reveals that longer trip distance and duration increase both requested-to-share and shared probabilities, but higher cost reduces the probability. The joint dependency of trip cost and distance also signifies a high level of nonlinearity, which cannot be captured by a simple per-mile cost variable. We also tested several other scenarios for model specifications such as substituting trip duration and cost by per-mile cost, removing trip duration, which is correlated with trip distance. However, the performance of all alternative scenarios was inferior to our preferred classifiers. Moreover, other tested classification models including logistic regression and support vector machine do not reach an acceptable recall likely because the relationship between high importance features cannot be explained as linear.

To ensure robustness of the model, we repeat the entire learning process for all candidate models using 100 bootstrapped random samples size of 8,000 from the whole dataset. To quantify the uncertainty in performance metrics, we measure accuracy, precision, and recall on the test sets. As Fig 8 shows, on average, the Ada model outperforms other models in all metrics for prediction of requested-to-share trips and GB has a slight edge for prediction of successfully shared trips. Overall, the narrow bounds of distributions of performance metrics ensures robustness to sampling and the training process in our classifications.

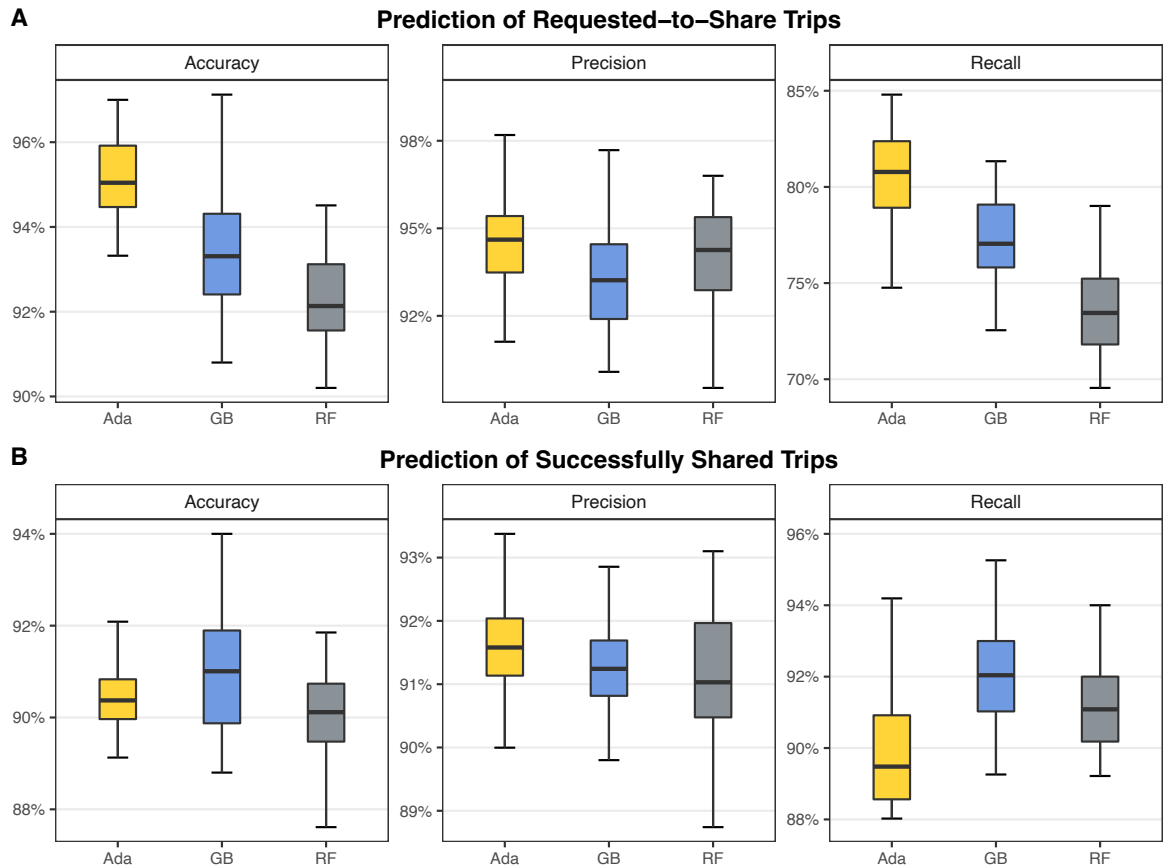

Fig 8. Distribution of model performance metrics over 100 bootstrapped random samples size of 8,000 from the entire dataset. The boxes describe 25th percentiles (left hinge), medians, and 75th percentiles (right hinge) and whiskers describe 1.5 times the interquartile range.

## 7. Discussion and Conclusion

In this study, we investigate sharing behavior in ride-hailing trips in Chicago. We show that WTS and successfully shared trips in Chicago halved throughout 2019 with a strictly declining trend, while the rate of successful matching, as well as trip volume and VMT, have been stable. A regression analysis reveals that a gradual increase in per-mile cost of the trip over time has been the major factor in the decline of WTS, especially for shorter trips which were likely substituted by solo trips. Using ensemble ML methods, we find that travel impedance variables (trip cost, distance, and duration) have the highest predictive power in predicting the propensity to share. Longer but less expensive requested-to-share trips are more likely to be successfully shared with another ride. A wide range of explanatory variables at the pickup and drop-off of trips, including socioeconomic, demographic, spatial and temporal attributes, built environment, and transit supply variables are loosely correlated with WTS and successful sharing, but do not bear significant predictive power (although the characteristics of all census tracts along the trip route could matter as well, we have not studied them here). WTS in EDAs (low incomes and high concentration of minorities) is nearly double that of non-EDAs. However, as shown in Fig. 3, since the EDA trips are longer and more expensive, lower price of shared trips has become a stronger factor encouraging riders to opt for sharing. In EDAs , the transit supply is also poor, suggesting that the convenience of door-to-door mobility combined with lower price of shared trips may be replacing active and public transit trips and come at the expense of expanding infrastructure and/or services to those areas (Barajas and Brown, 2021). This is consistent with Schaller (2021) which found sharing in ride-hailing is most popular in lieu of public transit and attracts passengers from existing shared modes (Schaller, 2021). Some studies, however, argue that ride-hailing and sharing to transit hubs have increased transit ridership (Hall et al., 2018).

Since travel impedance variables best explain sharing behavior compared to other variables, pricing signals have the highest potential to encourage riders to substitute solo rides with sharing. Thus, policymakers and TNCs can more efficiently allocate sharing incentives to increase the penetration of sharing and reduce congestion and environmental externalities of ride-hailing. Recognizing this, on January 6, 2020, the City of Chicago initiated a new congestion fee policy for ride-hailing trips with differential pricing for shared trips in the downtown area (The City of Chicago Business Affairs and Consumer Protection, 2019). The new policy increased the congestion tax on shared trips in the downtown congestion zone by 74% (0$.72/trip to $1.25/trip), but more than quadrupled the congestion tax on solo ride-hailing trips ($0.72/trip to $3/trip). Future research may assess the impacts of this new policy on sharing behavior.

In essence, our *trip-level* classification problem here is comparable to Hou et al. (Hou et al., 2020) and Xu et al. (Xu et al., 2021) in which the aggregate ratio of requested-to-share trips between O-D pairs at the census tract level is predicted as a regression-based problem. They found high mean squared error in prediction as well as high sensitivity of models to inclusion of O-D pairs with few connecting trips (outliers). Since travel impedance variables show little variance in O-D pair analysis aggregated at the census tract level, their predictive power is limited. Our classification models demonstrate high level of accuracy and robustness in direct prediction of shard-requested and successfully shared trips with the highest predictive powers in travel impedance variables. The association of factors with WTS does not necessarily guarantee causality. For instance, it is identified that the higher portion of black population or lower income in the pickup or drop-off census tracts are associated with greater WTS (Hou et al., 2020; Xu et al., 2021). However, cost, distance, and duration of trips are also higher in lower income and black-majority areas (EDAs) in Chicago (Fig. 3). Thus, a casual inference without constructing a structural model is likely unreliable.

There are several inherent limitations in the Chicago data as well as our approach of examining the sharing behavior and making inference. First and foremost, the Chicago dataset does not divulge details on characteristics of riders and trips. The choice of solo and shared for the rider is largely a function of fare difference between alternatives when the rider requests a ride. The data does not reveal the cost difference between alternatives choices and only report the final state of the trip. Future work can fill this gap by utilizing historical data between the same O-D pair, requested in the similar time of day, from different riders, as a proxy for the cost of alternative choices.

Rounding the fare to nearest $2.5 is a source of unknown bias in our inference, which prior studies have also acknowledged (Dean and Kockelman, 2021; Hou et al., 2020; Xu et al., 2021). Secondly, since the trips are anonymized, we cannot attribute the pattern to individual riders with specific sociodemographic attributes to identify the determinants of WTS. The data does not reveal wait time for requested-to-share rides versus solo rides, which is among the most important factors for WTS of individual trips. The suppression of pickup and drop-off locations to the census tracts creates large bias. The data does not carry any information on individual ridership characteristics, which hinders higher resolution identification of sharing preferences (for instance, the number of people in each requested ride or a unique id for each rider to assess the individual sharing behavior over multiple trips). Thus, in addition to the possibility of omitted variable bias in our inference, our findings cannot be generalized to human behavior in facing a choice and should not be interpreted as a choice experiment. Rather, our findings underscore the predictive power of trip-level attributes, which can still be leveraged to answer important questions associated with propensity for ride sharing. Finally, the sharing behavior after COVID-19 will likely change significantly. Since March 16, 2020, no shared ride was offered in Chicago, and it will likely continue

for a foreseeable future. Thus, it is important to revisit the data once shared rides are offered again and assess the post-pandemic behavior and how preference has changed.

## Code and Data Availability

All datasets used in this study are publicly available. The 2019 trip-level ride-hailing data can be accessed via https://data.cityofchicago.org/Transportation/TNC-2019-Trips/n4i4-nw6t. The data cleaning and visualizations were performed using R statistical software. The machine learning process is implemented using scikit-learn 0.23.2 in Python. All data and supporting codes are available from the authors upon a reasonable request.

## Appendix: Model training procedure

***Hyperparameter tuning.*** To ensure the model performance and address the potential risk of overfitting, we tune the models' hyperparameters. The regularization in the training process excludes the patterns that are unimportant for the prediction and do not generalize beyond the training set. These hyperparameters included core parameters and learning control parameters (e.g., learning rate and, number of trees, and maximum tree depth; see Table A-1 for all hyperparameters tuned).

Since we deal with unbalanced classifications (i.e., nearly 20% requested-to-share versus 80% solo trips and 70% successfully shared versus 30% unmatched trips), we maximize recall as the objective of hyperparameter tuning instead of prediction accuracy. The best set of parameters enables the model to generalize from the training set to the test set while maintaining the highest intended performance. We use randomized grid search for implementation of hyperparameter tuning to reduce the training time.

Table A-1. Hyperparameter tuning results using 5-fold cross-validation.

| Hyperparameter | Prediction of Requested-to-share | | | Prediction of Successfully Shared | | |
|---|---|---|---|---|---|---|
| | *Ada* | *GB* | *RF* | *Ada* | *GB* | *RF* |
| Number of Trees (number of estimators) | 100 | 100 | 50 | 100 | 100 | 150 |
| Max Depth | - | 7 | 5 | - | 5 | 5 |
| Number of Features | 10 | 5 | 8 | 8 | 8 | 5 |
| Learning Rate | 0.5 | 0.1 | - | 1 | 0.7 | - |
| Criterion | - | friedman_mse | gini | - | friedman_mse | entropy |

***Feature selection.*** Before confirming the final list of features to be included in the models, we perform feature selection, to identify the variables most relevant to the prediction and removing those that do not contribute to, or reduce, the predictive power of the model (Breiman, 2001). An incorrect generalization from an unintended property of the training set is called overfitting. The feature selection process is implemented trough permutation feature importance (Scikit-learn), which reports the relative importance of all variables as shown in Fig. 6 (normalized for comparison and only top 10 are shown). We exclude the variables that have an average impact on the prediction less than 0.5% when the variable value is randomly shuffled. This ensures that the 10 variables included in the final models (as listed in Fig. 6) are all stable and important for the prediction.

***Model validation.*** We validate models, a process to test the performance of the tuned and trained model on data different from the training set, that is, a validation set. This ensures that the models do not overfit, and performs well not only on the training set, but also out-of-sample. We use $k$-fold cross-validation, in which the data are split into $k$ folds and the model is fitted $k$ times, each time with a different fold chosen as a test set with the rest performing as a training set. We chose $k$ = 5 here.

***Model calibration.*** Our classification models return the probability of binary class (soft predictor). We calibrate the probabilistic prediction made by our model using the sigmoid approach (Zadrozny and Elkan, 2002). Calibration is essential for retrieving unbiased probability estimates from the model. The output of a well-calibrated model can be directly interpreted as an estimate of probability.

**_Performance metrics._** Accuracy, precision, and recall are compiled to assess the performance of models from different perspectives. All measures can be calculated from TP (true positives), TN (true negatives), FP (false positives), and FN (false negatives) of the confusion matrix. The accuracy is the proportion of correct predictions ($\frac{TP+TN}{n}$), the precision evaluates the fraction of correct classified instances among the ones classified as positive ($\frac{TP}{TP+FP}$), and the recall (sensitivity) quantifies the number of correct positive predictions made out of all positive predictions that could have been made ($\frac{TP}{TP+FN}$). For imbalanced classifications such as our problem, recall and precision are more important performance measures than accuracy. Unlike precision that only comments on the correct positive predictions out of all positive predictions, recall provides an indication of missed positive predictions (Hastie et al., 2009) and hence is a better metric for classification of requested-to-share and shared trips. The overall performance of a classifier can be evaluated by the area under the receiver operating characteristic curve (ROC-AUC), which is equivalent to the probability that the model will rank a randomly chosen positive instance higher than a randomly chosen negative instance (Hastie et al., 2009). Higher AUC reflects higher predictive performance of a model.

Supplementary Materials

# Sharing Behavior in Ride-hailing Trips: A Machine Learning Inference Approach

Morteza Taiebat, Elham Amini, Ming Xu

This PDF includes:

- Fig. S-1 – S-8

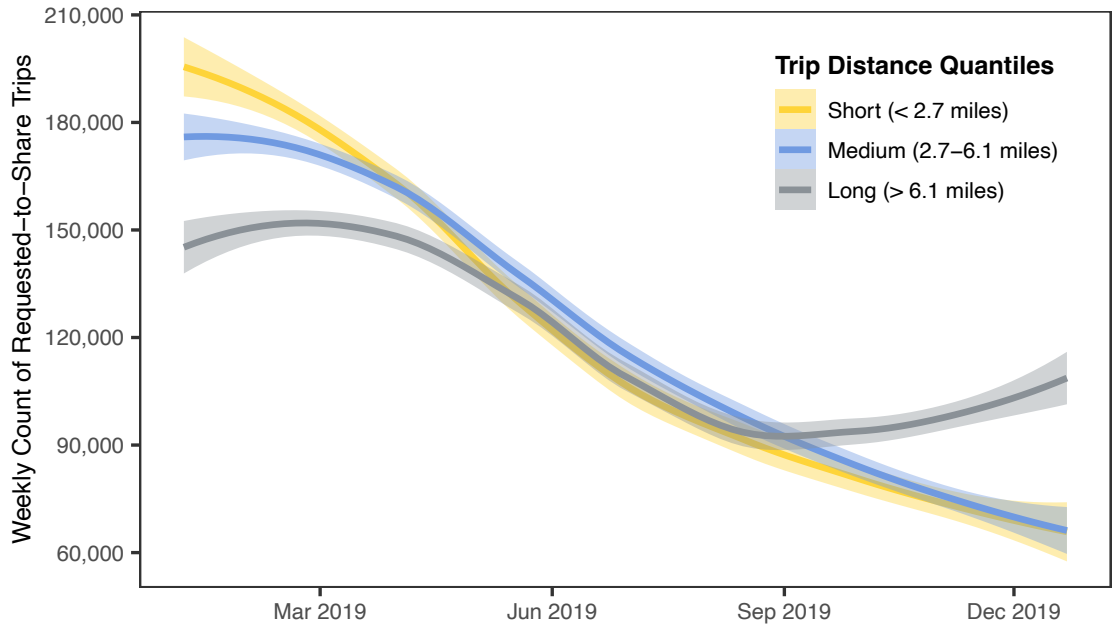

**Fig. S-1.** Smoothed trend of requested-to-share trip by trip distance. The requested-to-share trips are divided to three equally-sized quantiles: 0-33.4 percentile: short (less than 2.7 miles); 33.4-66.7 percentile: medium (2.7-6.1 miles); and 66.7-100 percentile: long (above 6.1 miles. Over time, the frequency of short and medium trips drops more long ones. Since the volume of trips did not statistically changed over time, it implies that the preference for shorter trips shifted to solo.

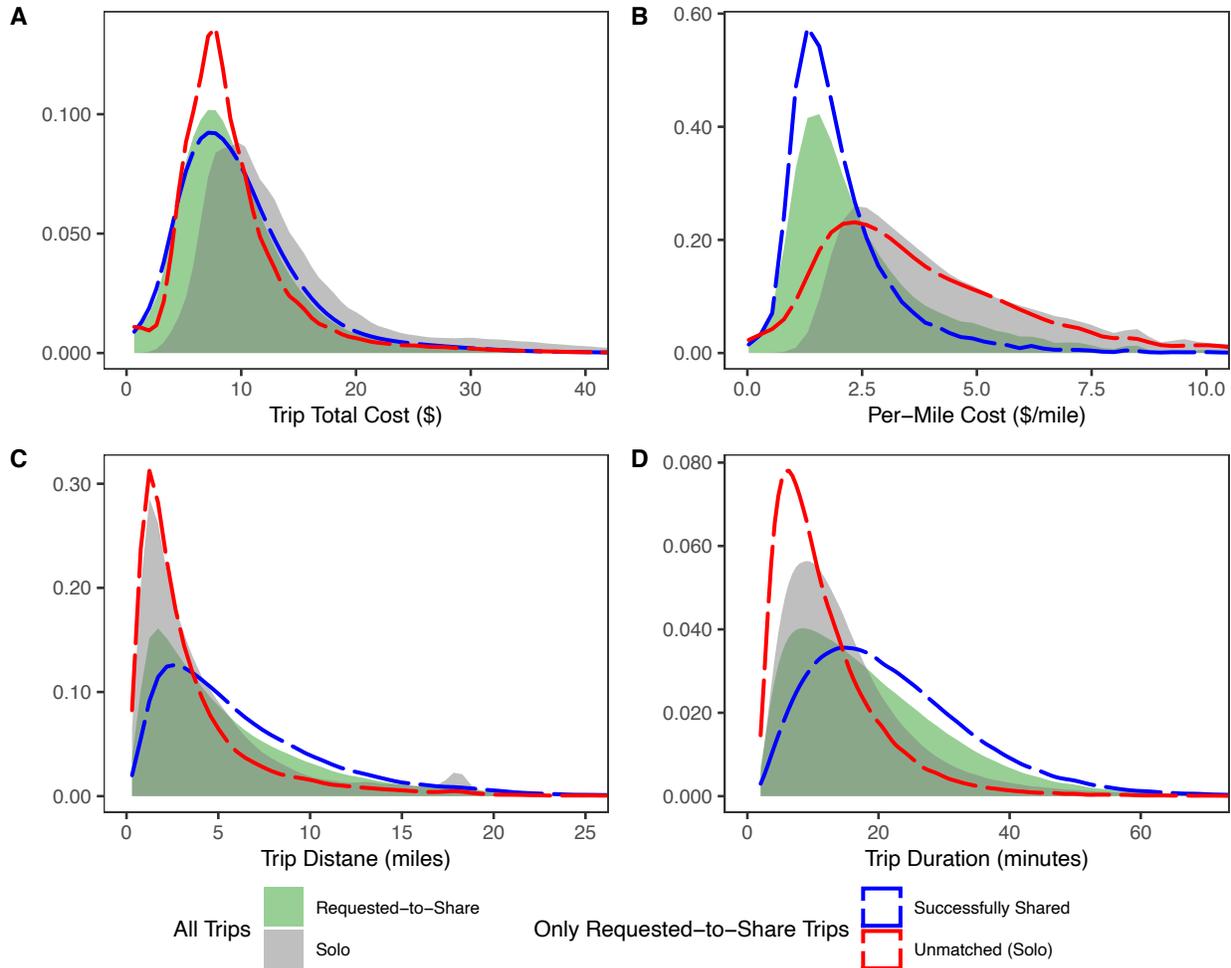

**Fig. S-2.** Kernel density distribution of key variables for requested-to-share trips, solo trips, shared trips, and unmatched requested-to-share trips. (A) trip total cost; (B) trip per-mile cost; (C) trip distance; (D) trip duration.

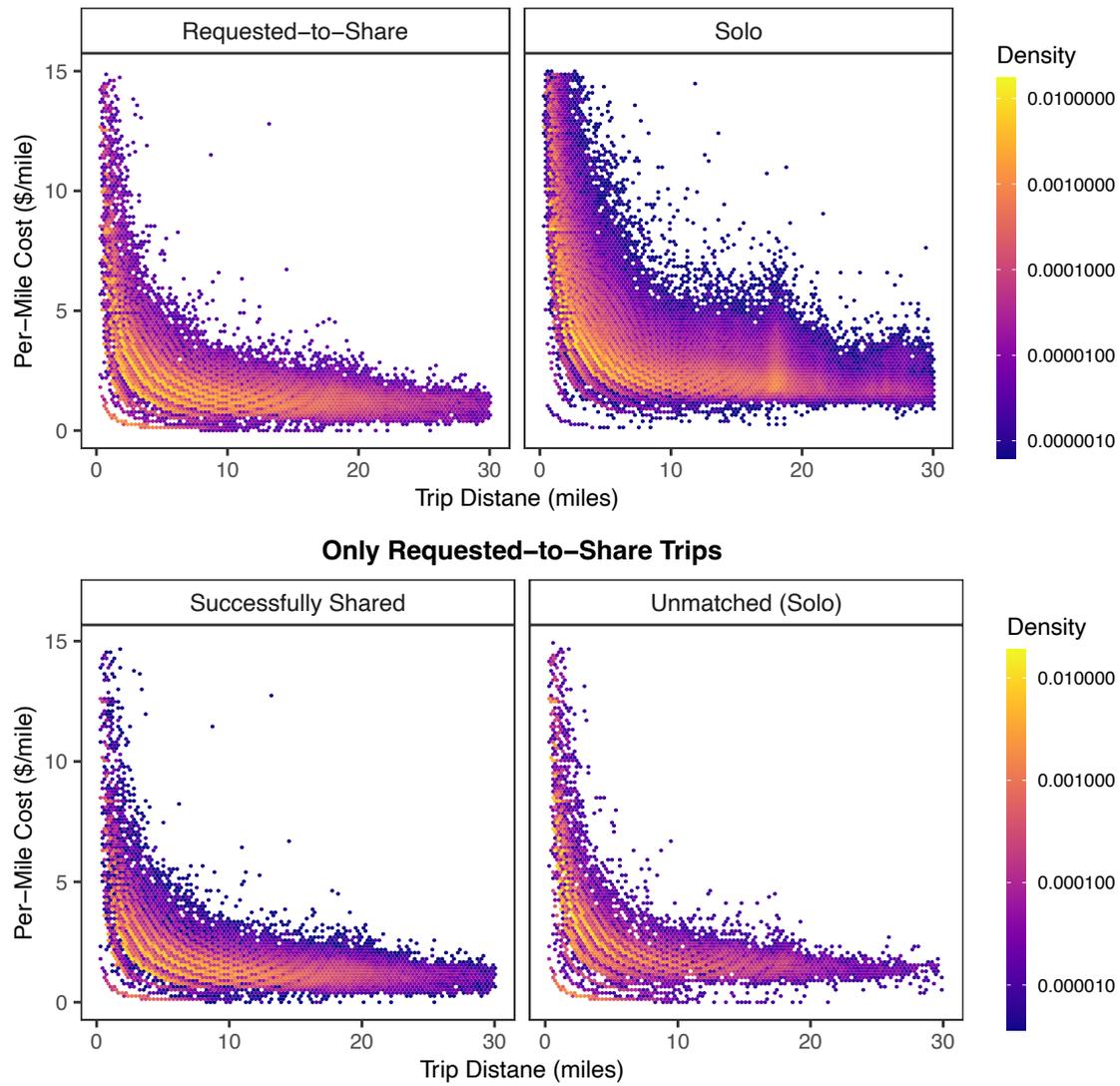

**Fig. S-3.** Joint density distribution of trip distance and per-mile cost for requested-to-share trips, solo trips, shared trips, and unmatched requested-to-share trips.

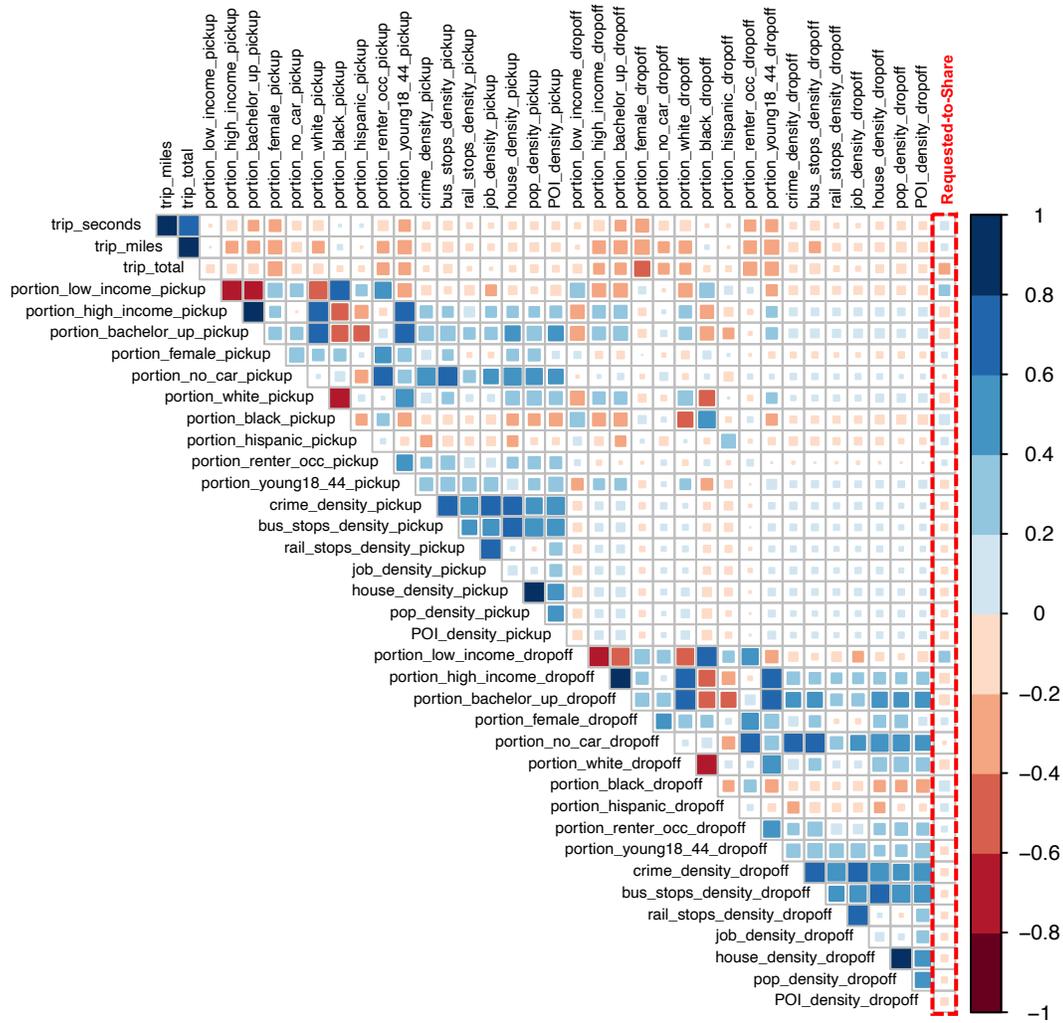

Fig. S-4. Correlation matrix between target response and explanatory variables. The target response is whether the trip is requested to be shared.

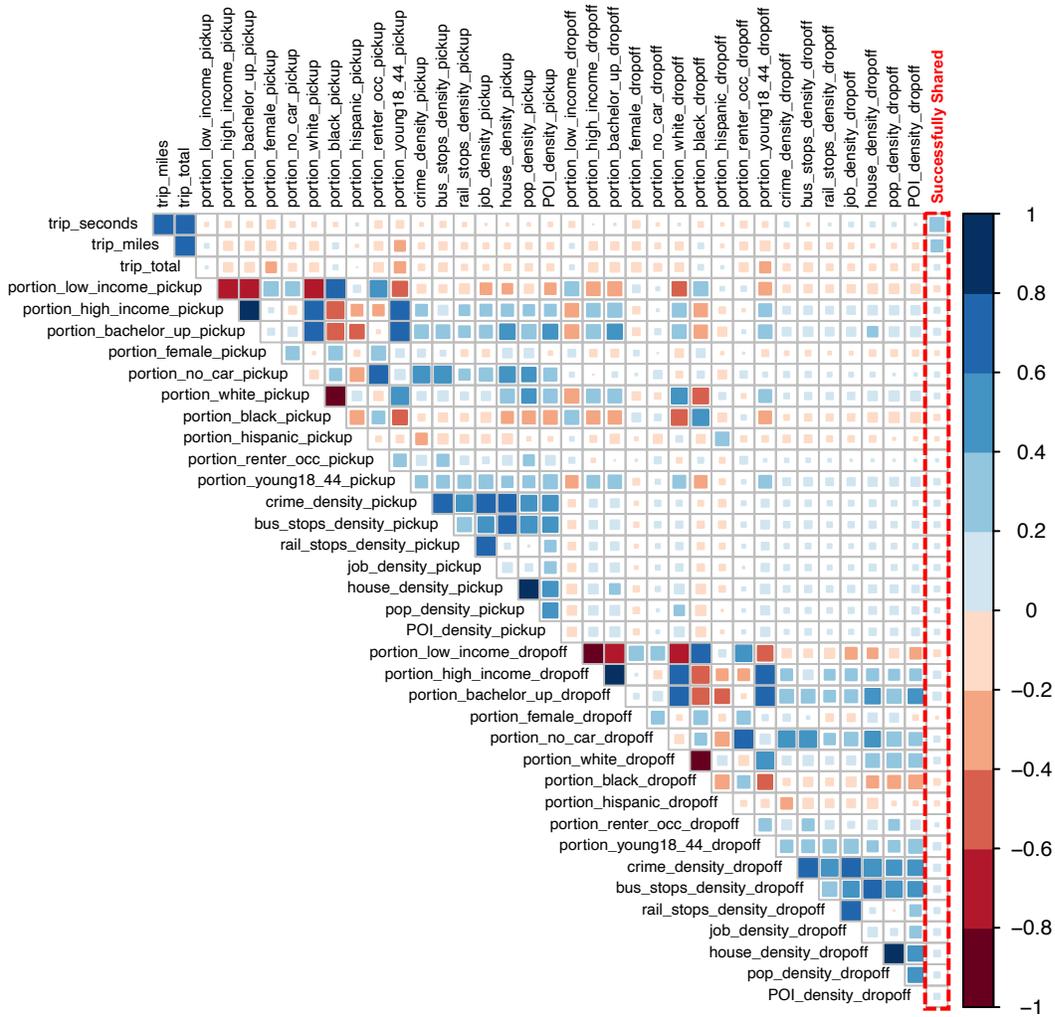

Fig. S-5. Correlation matrix between target response and explanatory variables. The target response is whether a requested-to-share trip is successfully shared or not.

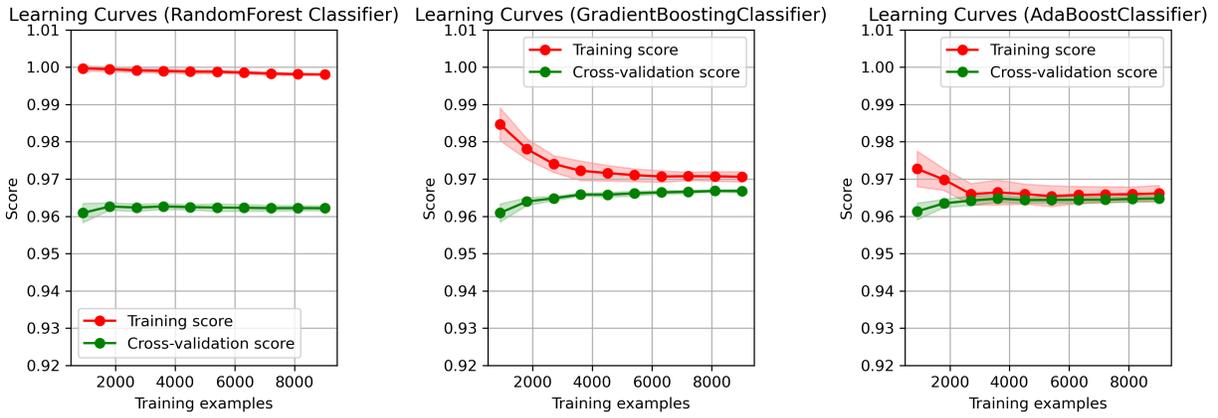

Fig. S-6. Learning curves with 5-fold cross validation for RF, GB, and Ada classifiers. A training sample of 6000 trips (8000 trips including test set) saturates all models with sufficient amount of data.

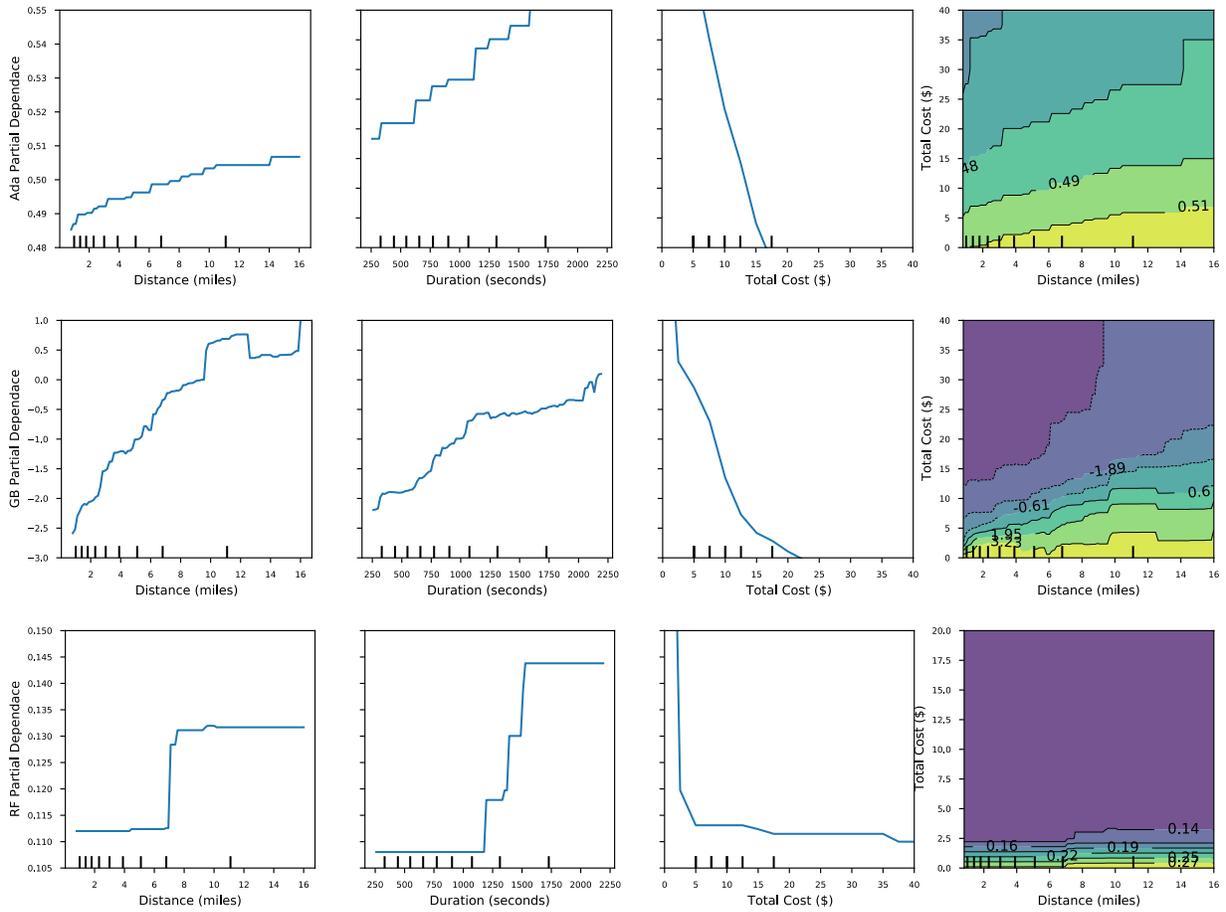

Fig. S-7. Partial dependence plots for three classification models of requested-to-share trips. Top three variables with highest permutation feature importance are shown. The partial dependence shows high nonlinearity between the target response and explanatory variables.

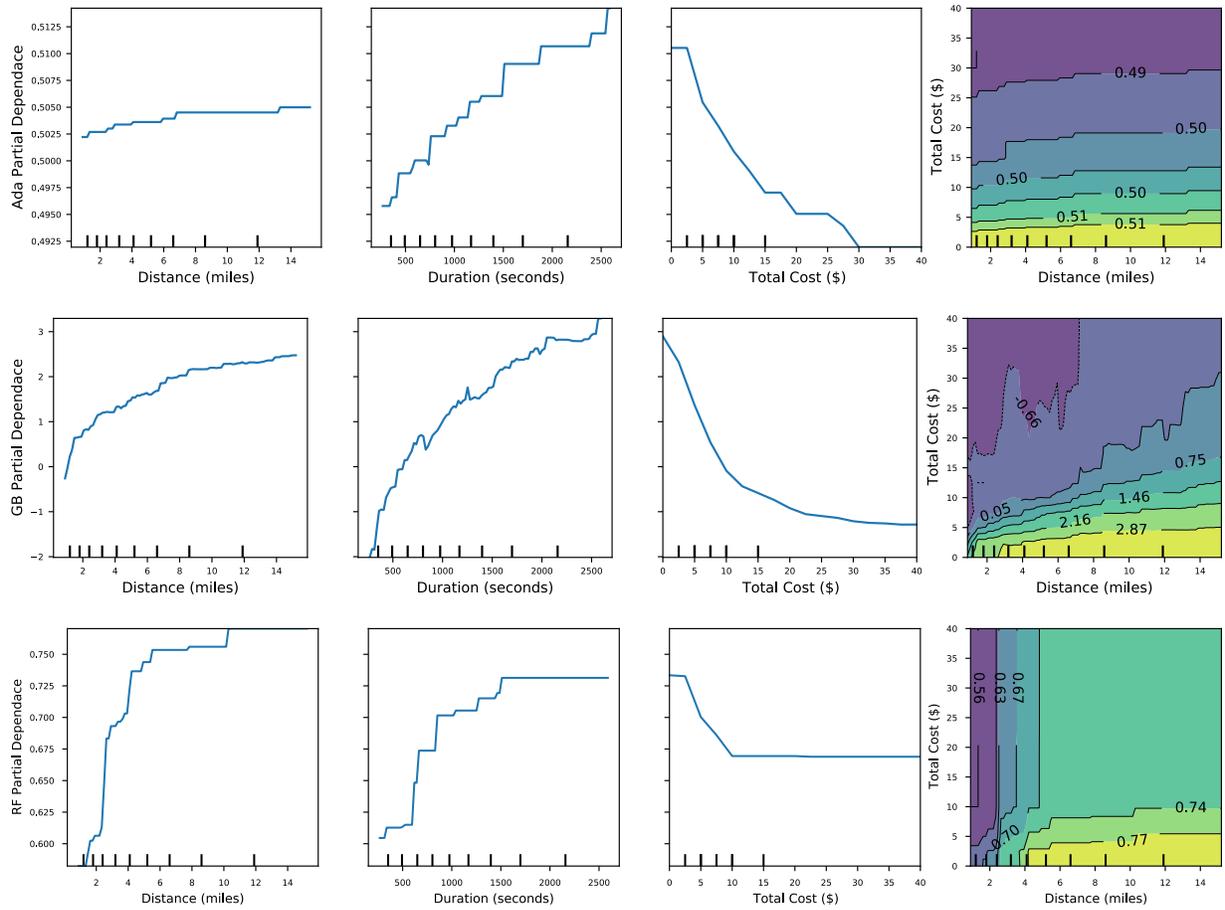

Fig. S-8. Partial dependence plots for three classification models of successfully shared trips. Top three variables with highest permutation feature importance are shown. The partial dependence shows high nonlinearity between the target response and explanatory variables.